\documentclass[10pt, a4paper]{article}
\usepackage{lrec2022} % this is the new LREC2022 Style
\usepackage{multibib}
\newcites{languageresource}{Language Resources}
\usepackage{graphicx}
\usepackage{tabularx}
\usepackage{soul}
\usepackage{adjustbox}
\usepackage{multirow}
\usepackage{amsmath}
\usepackage{titlesec}
\titleformat{\section}{\normalfont\large\bfseries\center}{\thesection.}{1em}{}
\titleformat{\subsection}{\normalfont\SmallTitleFont\bfseries\raggedright}{\thesubsection.}{1em}{}
\titleformat{\subsubsection}{\normalfont\normalsize\bfseries\raggedright}{\thesubsubsection.}{1em}{}
\renewcommand\thesection{\arabic{section}}
\renewcommand\thesubsection{\thesection.\arabic{subsection}}
\renewcommand\thesubsubsection{\thesubsection.\arabic{subsubsection}}
\usepackage{epstopdf}
\usepackage[utf8]{inputenc}

\usepackage{hyperref}
\usepackage{xstring}

\usepackage{color}

\usepackage{devanagari}
\usepackage{babel}

\usepackage{fancyhdr}
\title{EmoInHindi: A Multi-label Emotion and Intensity Annotated Dataset in Hindi for Emotion Recognition in Dialogues}

\name{Gopendra Vikram Singh*\thanks{* The authors are jointly the first authors},Priyanshu Priya*,Mauajama Firdaus*,Asif Ekbal,Pushpak Bhattacharyya}

\address{Department of Computer Science and Engineering \\
         Indian Institute of Technology Patna, Patna, India\\
        \{gopendra\_1921cs15, priyanshu\_2021cs26, mauajama.pcs16, asif, pb\}@iitp.ac.in\\}

\abstract{
The long-standing goal of Artificial Intelligence (AI) has been to create human-like conversational systems. Such systems should have the ability to develop an emotional connection with the users, %consequently, and 
hence emotion recognition in dialogues is an important task. %has gained popularity. 
Emotion detection in dialogues is a challenging task because humans usually convey multiple emotions with varying degrees of intensities in a single utterance. Moreover, emotion in an utterance of a dialogue may be dependent on previous utterances making the task more complex. 
Emotion recognition has always been in great demand.
However, most of the existing datasets for multi-label emotion and intensity detection in conversations are in English. To this end, we %propose 
create a large conversational dataset in Hindi named \textit{EmoInHindi} for multi-label emotion and intensity recognition in conversations containing 1,814 dialogues with a total of 44,247 utterances. We prepare our dataset in a Wizard-of-Oz manner for mental health and legal counselling of crime victims. Each utterance of the dialogue is annotated with one or more emotion categories from the 16 emotion classes including neutral, and their corresponding intensity values. We further propose strong contextual baselines that can detect emotion(s) and the corresponding intensity of an utterance given the conversational context. 
 \\ \newline  }
 
\begin{document}

\maketitleabstract
\thispagestyle{fancy}

\section{Introduction}
Emotions are fundamental human characteristics that have been researched for many years by researchers in psychology, sociology, medicine, computer science, and other domains. Ekman's six-class categorization \cite{ekman1992argument} and Plutchik's Wheel of Emotion which proposed eight basic bipolar emotions \cite{plutchik2013biological}, are two notable works in understanding and categorising human emotions. Emotions play an important role in our daily life and emotion detection in text has become a longstanding goal in Natural Language Processing (NLP). Emotions are inherently conveyed by messages in human communications. With the popularity of social media platforms like Facebook Messenger, WhatsApp and conversational agents like Amazon's Alexa, there is a growing demand for machines to interpret human emotions in real conversations for more personalized and human-like interactions.  %However, enabling machines to interpret and correctly understand emotions in human texts is difficult, partially because humans frequently rely on context and commonsense knowledge to communicate emotions, which machines find difficult to record.

The ability to effectively identify emotions in conversations is crucial for developing robust dialogue systems. There are two major types of dialogue systems: a task-oriented dialogue systems and a social (chit-chat) dialogue system. The former is concerned with creating a personal assistant capable of performing specific tasks, but the latter is concerned with capturing the conversation flow, which focuses more on the speaker's feelings. In both these systems, understanding
the user’s emotions is crucial for providing better user experience and maximizing the user satisfaction. Nowadays, %lots of
many websites, blogs, tweets, conversational agents support Hindi language and some of them use Hindi as a primary language as well. However, %maximum study on emotion detection in conversations has been motivated on English language
most studies of emotion in conversations have focused on English language interactions \cite{chen2018emotionlines,yeh2019interaction,hazarika2018conversational,ghosal2019dialoguegcn,kim2018attnconvnet,he2018joint,yu2018improving,huang2019seq2emo}; comparatively very little attention is given to emotion detection in regional languages like Hindi. Towards this end, we propose a novel conversational dataset \textit{EmoInHindi} for identifying emotions (e.g., joy, sad, angry, disgusted etc.) in textual conversations in Hindi language, where the emotion of an utterance is detected in the conversational context.

\subsection{Problem Definition}
\begin{figure*}[!ht]
    \centering
    \begin{adjustbox}{max width=0.6\linewidth}
    \includegraphics{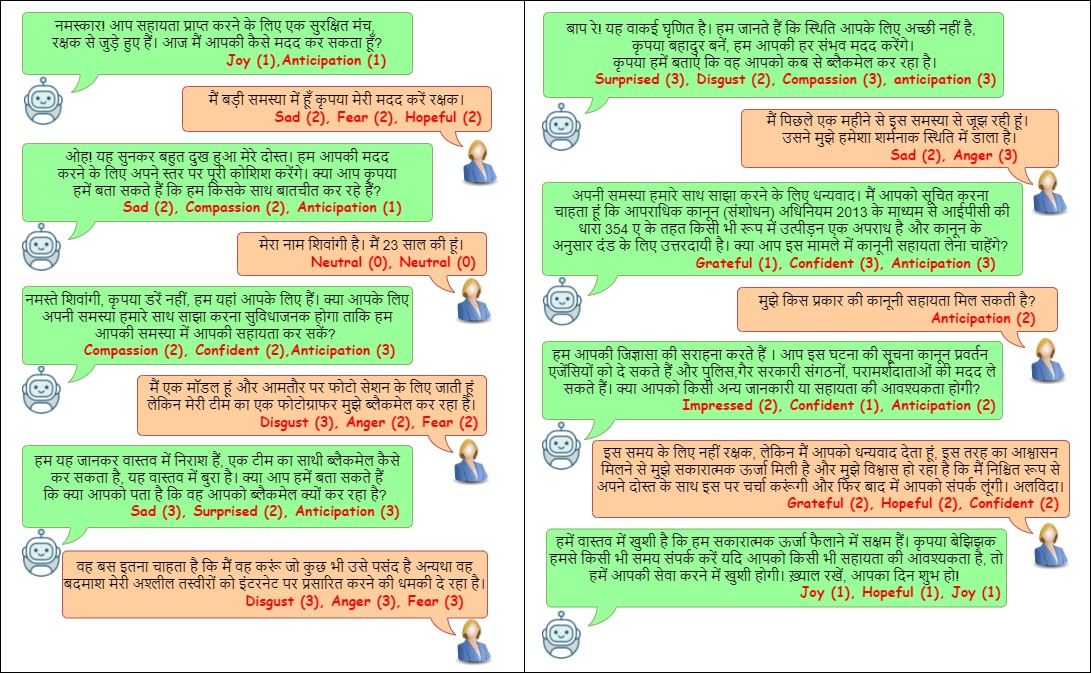}
    \end{adjustbox}
    \caption{Sample dialogue from our dataset with emotion and corresponding intensity annotation}
    \label{sampleDialog}
\end{figure*}

Given a textual utterance of a dialogue along with the conversation history (previous few utterances in dialogue), the task is to identify the emotion category(s) of each utterance from a set of pre-defined emotion categories and their corresponding intensity values. Formally, given the input utterance $U_t$ consisting of sequence of words $U_t = \{w_1,w_2,...,w_T\}$ and the conversation history $C$ consisting of sequence of utterances $C = \{U_1,U_2,...,U_{t-1}\}$, the task is to predict one or more emotion label, $e = \{e_1,e_2,...,e_L\}$ from $N$ pre-defined set of emotions and corresponding intensity value $i = \{i_1,i_2,...,i_L\}$, where $i_k \in \{0,1,2,3\}$ of the utterance $U_t$. Fig. \ref{sampleDialog} depicts a sample dialogue from our dataset, where each utterance is labeled with one or more underlying emotions and corresponding intensity value.

\subsection{Contribution}
The key contributions of our work are \textit{two-fold}:
\begin{itemize}
\setlength{\itemsep}{2pt}
\setlength{\parskip}{0pt}
\setlength{\parsep}{0pt}
    \item We propose \textit{EmoInHindi}%\footnote{https://www.ac.in/~ai-nlp-ml/resources}
    \footnote{https://www.iitp.ac.in/~ai-nlp-ml/resources.html}, the currently largest Hindi conversational dataset labeled with multiple emotions and their corresponding intensity values.
    \item We setup strong baselines for utterance-level multiple emotion and intensity detection task and report their results for identifying emotion(s) and the corresponding intensity expressed in an utterance of a dialogue written in Hindi.
\end{itemize}

\section{Related Work}
With the development in Artificial Intelligence (AI), emotion classification has become a significant task because of its importance in many downstream tasks like response generation for conversational agents, customer behavior modeling, multimodal interactions and many more. Recently, \newcite{kim2018attnconvnet,he2018joint,yu2018improving,huang2019seq2emo} investigated multi-label emotion classification for textual data. \newcite{kim2018attnconvnet} performed multi-label emotion classification on twitter data using multiple Convolution Neural Network (CNN) networks along with self-attention and \newcite{huang2019seq2emo} employed sequence-to-sequence framework for multi-label emotion classification. \newcite{yu2018improving} improved the performance of multi-label emotion classification on twitter data by using transfer learning. Our present study differs from the previous multi-label emotion classification research in that we categorise emotions of utterances of conversations, which require contextual knowledge from previous utterances, making the task more difficult and intriguing.\\\\Recently, emotion recognition in conversations \cite{chen2018emotionlines,yeh2019interaction,hazarika2018conversational,ghosal2019dialoguegcn,firdaus2020meisd} has been in demand. \newcite{li2017dailydialog} developed a high-quality multi-turn dialogue dataset, DailyDialog labelled with emotion information.  \newcite{chen2018emotionlines} proposed a corpus named EmotionLines for detecting emotions in dialogues gathered from Friends TV scripts and private Facebook messenger dialogues. EmoContext \cite{chatterjee2019semeval} is the another publicly available conversational dataset for emotion detection. All these datasets mostly focus on chit-chat dialogues. Lately, \newcite{feng2021emowoz} introduced EmoWOZ, a large-scale manually emotion-annotated corpus of task-oriented
dialogues. \\\\Most of the existing methods and resources developed for emotion analysis %in conversations
are available in English \cite{yadollahi2017current}. Lately, there has been work on developing resources for detecting emotions from Hindi text. For instance, \newcite{vijay2018corpus} created a corpus, consisting of sentences from Hindi-English code switched language used in social media for predicting emotions. Likewise, \newcite{koolagudi2011iitkgp} proposed a Hindi corpus consisting of sentences taken from auditory speech signals for emotion analysis task. Another Hindi dataset consisting of sentences from news documents of disaster domain for emotion detection was proposed by \newcite{ahmad2020borrow}. \newcite{kumar2019bhaav} presented a largest annotated corpus in Hindi comprising of sentences taken from various short stories in which each sentence is annotated with relevant emotion categories given the context of a sentence. However, all of these works are focused on non-conversational settings. The long-term goal of our present work is to build a dialogue system capable of having a conversation with the user in Hindi. Such a system should not only be able to respond in Hindi according to the user’s intent, but its utterances should also be aligned with the user’s emotional state. As opposed to existing works on emotion detection from textual data in Hindi language, our present work provides a multi-label emotion and intensity annotated conversational dataset in Hindi. % for the classification of multiple emotions and corresponding intensity in the given utterance of a dialogue. 

\section{Dataset}
In this section, we describe the complete details of our EmoInHindi dataset.

\subsection{Dataset Preparation}
The dataset that we prepared for our experiment comprises of dialogues focused on mental health counselling and legal assistance for women and children who have been victims of various types of crimes ranging from domestic violence, workplace harassment, matrimonial fraud, to cybercrimes like cyber stalking, online harassment, masquerading and trolling. We construct the dataset in Hindi in Wizard-of-Oz \cite{kelley1984iterative} style. Every dialogue in the dataset starts with a basic description of the victim, after which the victim is asked about the problem and accordingly provided with the required assistance. The crime victims need emotional comfort and support for expressing their feelings freely, hence the dialogue systems should interact with the users empathetically. Such conversational agents that comprehend human emotions assist in enhancing the user's communication with the system, thereby strengthening the communication in a positive direction \cite{martinovsky2006error,prendinger2005empathic}. We have annotated every utterance in a dialogue with multiple appropriate emotion categories and their corresponding intensity. 

%%%%%%%%%%%%%%%%%%%%%%%%%%%%%

\begin{table*}[ht]
    %\footnotesize
    \begin{center}
    \begin{adjustbox}{max width=1.1\linewidth}
    \begin{tabular}{|lll|}
    \hline
    \multicolumn{1}{|l|}{\textbf{Speaker}} &
      \multicolumn{1}{l|}{\textbf{Utterances}} &
      \textbf{Emotion \& corresponding intensity} \\ \hline
    \multicolumn{3}{|c|}{\textit{\textbf{Dialogue 1}}} \\ \hline
    \multicolumn{1}{|l|}{\textit{\textbf{Victim}}} &
      \multicolumn{1}{l|}{\begin{tabular}[c]{@{}l@{}} 
      %\begin{sanskrit}मैं ठगा गया हूँ। \end{sanskrit}
      \def\DevnagVersion{2.17}{\dn m\4{\qva} WgA gyA \8{h}\1.}
      \\ (I am cheated.)\end{tabular}}  &
      sad (1) \\ \hline
    \multicolumn{1}{|l|}{\textit{\textbf{Agent}}} &
      \multicolumn{1}{l|}{\begin{tabular}[c]{@{}l@{}} %\begin{sanskrit} मुझे यह सुनकर दुख हुआ मेरे दोस्त। क्या आप कृपया मुझे बता सकते हैं कि आपको किसने धोखा दिया है?\end{sanskrit}
      \def\DevnagVersion{2.17}{\dn \7{m}J\? yh \7{s}nkr \7{d}K \7{h}aA m\?r\? do-t. \3C8wA aAp \9{k}pyA \7{m}J\? btA skt\? h\4{\qva} Ek aApko Eksn\? DoKA EdyA h\4{\rs ?\re}}
      \\ (I am sorry to hear this my friend. Could you please let me know who has cheated you?)\end{tabular}} &
      sad (1), anticipation (1) \\ \hline
    \multicolumn{1}{|l|}{\textit{\textbf{Victim}}} &
      \multicolumn{1}{l|}{\begin{tabular}[c]{@{}l@{}}%\begin{sanskrit} मुझे मेरे साथी ने सिर्फ संपत्ति के लिए धोखा दिया है।\end{sanskrit}
      \def\DevnagVersion{2.17}{\dn \7{m}J\? m\?r\? sATF n\? EsP\0 s\2pE\381w k\? Ele DoKA EdyA h\4.}
      \\ (I am cheated by my partner just for the sake of property.)\end{tabular}} &
      sad (1) \\ \hline
    \multicolumn{1}{|l|}{\textit{\textbf{Agent}}} &
      \multicolumn{1}{l|}{\begin{tabular}[c]{@{}l@{}} %\begin{sanskrit}मैं समझता हूं कि स्थिति आपके लिए अच्छी नहीं है। यह बहुत अच्छा होगा यदि आप इस पर कुछ और जानकारी साझा कर सकें ताकि हम आपकी बेहतर सहायता कर सकें।  \end{sanskrit}
      \def\DevnagVersion{2.17}{\dn m\4{\qva} smJtA \8{h}\2 Ek E-TEt aApk\? Ele aQCF nhF{\qva} h\4. yh b\7{h}t aQCA hogA yEd aAp is pr \7{k}C aOr jAnkArF sAJA kr sk\?{\qva} tAEk hm aApkF b\?htr shAytA kr sk\?{\qva}. }
      \\ (I understand the situation is not good for you. It would be great if  you could share few more information on this so that \\we could better assist you.)\end{tabular}} &
      compassion (1), anticipation (2) \\ \hline
    \multicolumn{3}{|c|}{\textit{\textbf{Dialogue 2}}} \\ \hline
    \multicolumn{1}{|l|}{\textit{\textbf{Victim}}} &
      \multicolumn{1}{l|}{\begin{tabular}[c]{@{}l@{}} %\begin{sanskrit}मैं ठगा गया हूँ। \end{sanskrit}
      \def\DevnagVersion{2.17}{\dn m\4{\qva} WgA gyA \8{h}\1.}
      \\ (I am cheated.)\end{tabular}} &
      sad (3) \\ \hline
    \multicolumn{1}{|l|}{\textit{\textbf{Agent}}} &
      \multicolumn{1}{l|}{\begin{tabular}[c]{@{}l@{}} %\begin{sanskrit} यह सुनकर वास्तव में निराशा हुई, मेरे प्रिय। क्या आप यह साझा करना चाहेंगे कि आपको किसने धोखा दिया है?\end{sanskrit}
      \def\DevnagVersion{2.17}{\dn yh \7{s}nkr vA-tv m\?{\qva} EnrAfA \7{h}I{\rs ,\re} m\?r\? E\3FEwy. \3C8wA aAp yh sAJA krnA cAh\?{\qva}g\? Ek aApko Eksn\? DoKA EdyA h\4{\rs ?\re}}
      \\ (This is really disappointing to hear, my dear. Would you mind sharing who has cheated you?)\end{tabular}} &
      sad (2), anticipation (1) \\ \hline
    \multicolumn{1}{|l|}{\textit{\textbf{Victim}}} &
      \multicolumn{1}{l|}{\begin{tabular}[c]{@{}l@{}} %\begin{sanskrit} मेरे पति ने मुझे दूसरी औरत के लिए धोखा दिया। वह एक गिरा हुआ इंसान है।\end{sanskrit}
      \def\DevnagVersion{2.17}{\dn m\?r\? pEt n\? \7{m}J\? \8{d}srF aOrt k\? Ele DoKA EdyA. vh ek EgrA \7{h}aA i\2sAn h\4.}
      \\ (My husband cheated on me for another women. He is such a creep.)\end{tabular}} &
      sad (3), anger (3) \\ \hline
    \multicolumn{1}{|l|}{\textit{\textbf{Agent}}} &
      \multicolumn{1}{l|}{\begin{tabular}[c]{@{}l@{}} %\begin{sanskrit}हम आपका दर्द समझते हैं। कृपया शांत हो जाएं; हम आपके साथ हैं और आपकी हर संभव मदद करने की पूरी कोशिश करेंगे। यदि आप सहज हैं, तो बेहतर सहायता के लिए हम आपसे कुछ और विवरण पूछना चाहेंगे। \end{sanskrit}
      \def\DevnagVersion{2.17}{\dn hm aApkA dd\0 smJt\? h\4{\qva}. \9{k}pyA fA\2t ho jAe\2{\rs ;\re} hm aApk\? sAT h\4{\qva} aOr aApkF hr s\2Bv mdd krn\? kF \8{p}rF koEff kr\?{\qva}g\?. yEd aAp shj h\4{\qva}{\rs ,\re} to b\?htr shAytA k\? Ele hm aAps\? \7{k}C aOr EvvrZ \8{p}CnA cAh\?{\qva}g\?. }
      \\ (We understand your pain. Please calm down; we are with you and will do our best to help you in every possible way. If you are comfortable, we would like to ask \\you few more details for better assistance.)\end{tabular}} &
      \begin{tabular}[c]{@{}l@{}} compassion (2), compassion (3),\\ anticipation (2) \end{tabular}\\ \hline
    \end{tabular}
    \end{adjustbox}
    \caption{Few utterances of dialogues with customized empathetic responses}
    \label{tab:sampledialogue}
    \end{center}
    \end{table*}
    
%%%%%%%%%%%%%%%%%%%%%%%%%%%%%%%%%%%%%%%%%%%%%%%%%%
   
\subsection{Guidelines for Dataset Preparation}
We contacted an expert in mental health counselling from our institute health department % National Institute of Mental Health and Neurosciences (NIMHANS)\footnote{https://nimhans.ac.in/}, Bengaluru so that we could 
 to understand the flow of dialogues in victims' situations to create the conversations. % and we drafted guidelines for the dialogue creation for our experiment that are as follows: 
At first, we tried to find out the problems of the victims and assessed their psychological needs. While counselling the victims, we make sure to be patient and kind towards them. The victims were provided a non-judgemental environment to share as much information they are comfortable to and if the victims decide to report the assault, seek medical attention, or contact organizations that can help them, then we assist them accordingly by providing the relevant legal, medical and organization information. Eventually, this was created to help the victims identify ways in which the victims can re-establish their sense of physical and emotional safety and provided a few basic safety suggestions to them so that they are aware of the crimes and can prevent such events in the future.

\subsection{Annotation}
The utterances in every dialogue of our proposed Hindi dataset is annotated with one or more appropriate emotion categories and their corresponding intensity. For annotating the dataset, we consider 15 emotions, namely \textit{Anticipation, Confident, Hopeful, Anger, Sad, Joy, Compassion, Fear, Disgusted, Annoyed, Grateful, Impressed, Apprehensive, Surprised, Guilty} as emotion labels for the utterances in a dialogue. The emotion annotation list has been extended to incorporate one more label, namely \textit{Neutral}. The \textit{“Neutral”} label is designated to utterances having no-emotion. While annotating the dataset, every utterance in a given dialogue is labeled with one or more emotions. Every emotion label is accompanied with an intensity value ranging from 1-3, with 1 indicating the lower intensity and 3 the highest. The \textit{Neutral} label has intensity value of 0. 
\\\\For annotating the utterances in our dataset, we employ three annotators highly proficient in Hindi and have prior experience in labelling emotions in conversational settings. The guidelines for annotation along with some examples were explained to the annotators before starting the annotation process. The annotators were asked to label each utterance of every dialogue with emotion(s) and corresponding intensity value. We achieve the overall Fleiss’ \cite{fleiss1971measuring} kappa score of 0.84 for the emotions, 0.88 for intensity, which can be considered reliable. To determine the final label of the utterances, we use majority vote.
%%%%%%%%%%%%%%%%%%%%%%%%%%%%%%%%%%%%%%%%%%%%%
\begin{figure}[!ht]
    \centering
    \begin{adjustbox}{max width=0.8\linewidth}
    \includegraphics{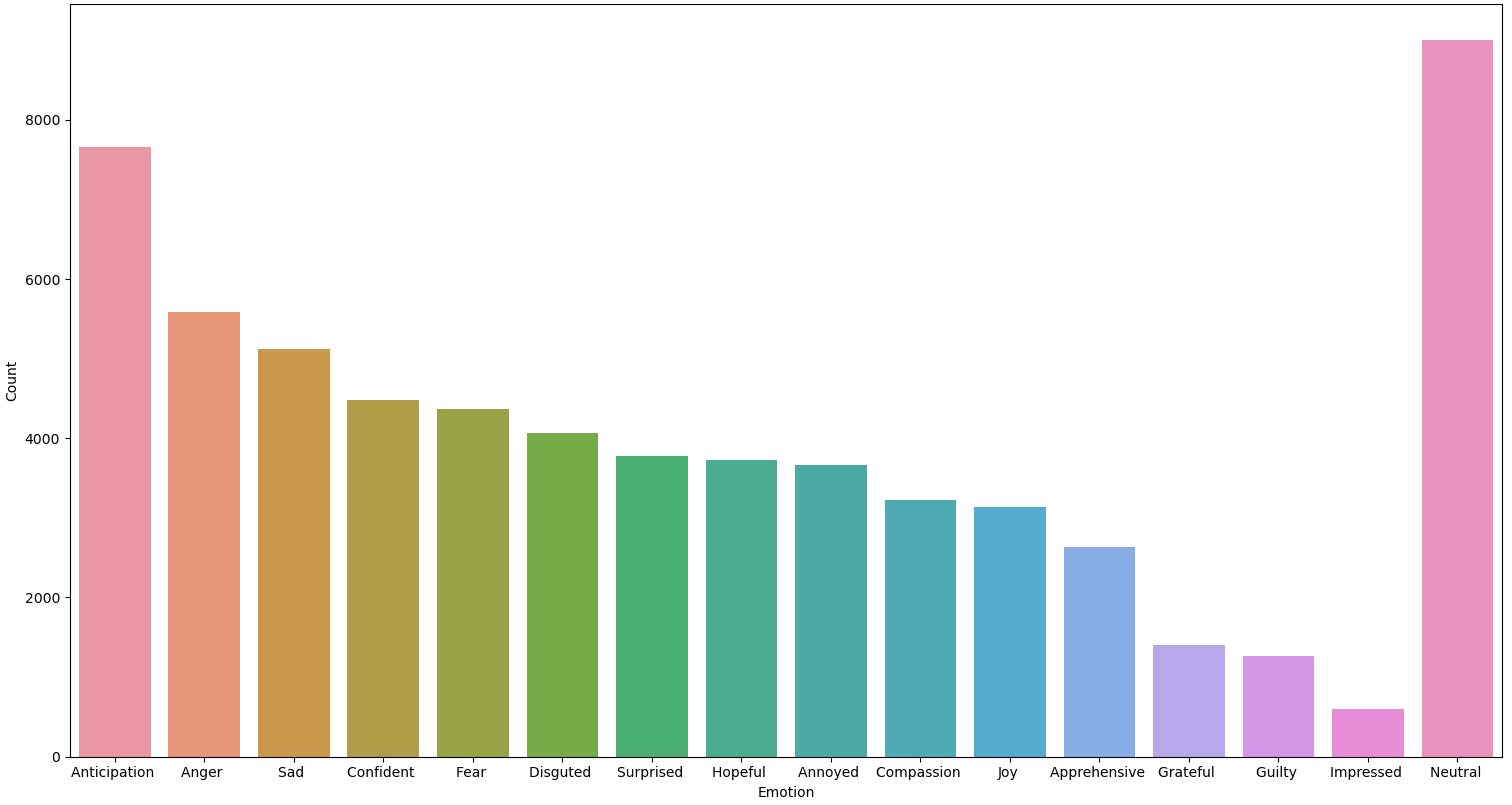}
    \end{adjustbox}
    \caption{Distribution of emotions in EmoInHindi dataset}
    \label{arch}
\end{figure}
%%%%%%%%%%%%%%%%%%%%%%%%%%%%%%%%%%%%%%%%%%%%%%%

\subsection{Challenges}
\textbf{Generic Challenges:} Counselling the victim and providing relevant assistance to them is a challenging task. If the intelligent agent does not seem supportive and understanding, the victim feels even more frightened and alone. Consequently, we came across various challenges while creating our conversational dataset which are as follows:
\begin{itemize}
\setlength{\itemsep}{2pt}
\setlength{\parskip}{0pt}
\setlength{\parsep}{0pt}
    \item  Counselling the victims according to their needs and mental state was a difficult assignment because of distinct mental state and need of every single individual. 
    \item Difficult to recognize and customize empathetic messages for different individuals as each individual has different emotional state. Depending upon the situation, one individual may show few emotions with lower intensity while the other may express his/her emotions intensely. Hence, replying empathetically according to the user's state is crucial for creating an amicable environment for them. For example, consider the following few utterances of two different dialogues from our dataset with emotion(s) and corresponding intensity shown in parentheses in Table \ref{tab:sampledialogue} which shows different empathetic responses of the agent depending upon situation and emotional state of different users.
    
    % \textbf{Dialogue 1}
    % \\\\\textbf{\textit{Victim}}: \textit{I am cheated.} (sad)(1)
    % \\\textbf{\textit{Agent}}: \textit{I am sorry to hear this my friend. Could you please let me know who has cheated you?} (sad, anticipation)(1,1)
    % \\\textbf{\textit{Victim}}: \textit{I am cheated by my husband just for the sake of property.} (sad)(1)
    % \\\textbf{\textit{Agent}}: \textit{I understand the situation is not good for you. It would be great if you could share few more information on this so that we could better assist you.} (compassion, anticipation)(1,2)
    % \\\\\textbf{Dialogue 2}
    % \\\\\textbf{\textit{Victim}}: \textit{I am cheated.} (sad)(3)
    % \\\textbf{\textit{Agent}}: \textit{This is really disappointing to hear, my dear. Would you mind sharing who has cheated you?} (sad, anticipation)(2,1)
    % \\\textbf{\textit{Victim}}: \textit{My husband cheated on me for another women. He is such a creep.} (sad, anger)(3,3)
    % \\\textbf{\textit{Agent}}: \textit{We understand your pain. Please calm down; we are with you and will do our best to help you in every possible way. If you are comfortable, we would like to ask you few more details for better assistance.} (compassion, compassion, anticipation)(2, 3, 2)
    \item Providing relevant and appropriate legal information to the victims.
    \item Providing step-by-step guidelines to victims for reporting the assault to the law enforcement agencies.
    % \item Annotating the data with appropriate emotion and corresponding intensity. The system needs to capture the correct emotion and accordingly handle the user by replying emapthetically. For example, if the user is angry, the system needs to pacify; if the user is furious about something, then the system should console the victim as per his/her emotional state.
    \item Helping the victim in re-establishing their sense of physical and emotional safety by being empathetic towards them.
\end{itemize}

\begin{table}[!ht]
\begin{center}
\begin{adjustbox}{max width=0.8\linewidth}
\begin{tabular}{|l|l|}
\hline
\textbf{Metrics}& \textbf{Dataset} \\ \hline
\textit{\# Dialogues}   & 1814   \\ \hline
\textit{\# Utterances}   & 44247    \\ \hline
\textit{Avg. utterances per dialogue}  & 24.39      \\ \hline
\textit{Avg. \# of emotions per dialogue}  & 1.41    \\ \hline
\textit{Avg. \# of emotions per utterance}  & 1.43      \\ \hline
\textit{\# of unique tokens}   & 7036\\ \hline
\textit{Avg. \# of tokens per utterance}   & 18.67      \\ \hline
\end{tabular}
\end{adjustbox}
\caption{Dataset statistics}
\label{datastat}
 \end{center}
\end{table}

\begin{table}[!ht]
\begin{center}
\begin{adjustbox}{max width=0.7\linewidth}
\begin{tabular}{|l|l|}
\hline
\textbf{Emotions}     & \textbf{Dataset}  \\ \hline
Anticipation & 7654   \\ \hline
Anger    & 5582    \\ \hline
Sad      & 5118   \\ \hline
Confident        & 4477    \\ \hline
Fear          & 4368       \\ \hline
Disguted          & 4060      \\ \hline
Surprised      & 3778    \\ \hline
Hopeful         & 3729   \\ \hline
Annoyed    & 3660    \\ \hline
Compassion      & 3218  \\ \hline
Joy     & 3130    \\ \hline
Apprehensive    & 2637  \\ \hline
Grateful & 1406        \\ \hline
Guilty     & 1269    \\ \hline
Impressed       & 595  \\ \hline
Neutral      & 9003    \\ \hline
\end{tabular}
\end{adjustbox}
\caption{Emotion Distribution}
\label{emotionstat}
 \end{center}
\end{table}

\begin{table*}[ht]
\begin{center}
\begin{adjustbox}{max width=1.1\linewidth}
\begin{tabular}{|l|l|l|l|l|l|l|}
\hline
\textbf{Dataset} &
  \textbf{\# of sentences} &
  \textbf{\# of emotions} &
  \textbf{\begin{tabular}[c]{@{}l@{}}Emotion Intensity\\ Annotation\end{tabular}} &
  \textbf{Language} &
  \textbf{Conversational} &
  \textbf{Multi-label} \\ \hline
\cite{vijay2018corpus} & 2866 & 6 & No
&  \begin{tabular}[c]{@{}l@{}}Hindi-English\\ code-mixed\end{tabular} &
  No  & No\\ \hline
\cite{koolagudi2011iitkgp}    & 12000 & 8  & No & Hindi & No & No \\ \hline
\cite{harikrishna2016emotion} & 780   & 5  & No & Hindi & No & No\\ \hline
\cite{kumar2019bhaav}         & 20304 & 5  & No & Hindi & No & No \\ \hline
\cite{ahmad2020borrow}        & 2668  & 9  & No & Hindi & No & No \\ \hline
%Proposed Dataset (Dataset) 
EmoInHindi & 44247      & 16 & Yes & Hindi & Yes & Yes \\ \hline
\end{tabular}
\end{adjustbox}
\caption{Comparison of different datasets and our proposed \textit{EmoInHindi} dataset %NC and C denotes non-conversational and conversational dataset settings, respectively. SL and ML denotes the Single-Label and Multi-Label emotion annotation, respectively
}
\label{compdata}
\end{center}
\end{table*}

\textbf{Annotation Challenges:}
The system needs to capture the correct emotion and accordingly handle the user by replying empathetically. Annotating the data with appropriate emotion and corresponding intensity is sometimes challenging. Apart from generic challenges mentioned in the previous section, we came across a few challenges while annotating the emotions which are as follows:
\begin{itemize}
\setlength{\itemsep}{2pt}
\setlength{\parskip}{0pt}
\setlength{\parsep}{0pt}
    \item \textbf{Identification of implicit emotions:} It is not always the case that emotions are communicated explicitly. We asked our annotators to identify both explicit as well as implicit emotions in the utterances. An example of explicitly expressed emotion would be \textit{Example 1} in which the speaker is clearly expressing that she is sad through the words %\begin{sanskrit}नर्क\end{sanskrit}
    \def\DevnagVersion{2.17}{\dn nk\0}(hell) ,%\begin{sanskrit}बुरा\end{sanskrit}
    \def\DevnagVersion{2.17}{\dn \7{b}rA}(bad) and %\begin{sanskrit}थक\end{sanskrit}
    \def\DevnagVersion{2.17}{\dn Tk}(tired) because her husband tortures her. 
    \\\\\textit{Example 1}: %\begin{sanskrit}{मेरी जिंदगी नर्क हो गई है, मुझे आजकल बहुत बुरा लग रहा है। मैं अपने पति और उसकी यातनाओं से थक चुकी हूं ।}\end{sanskrit}
    \def\DevnagVersion{2.17}{\dn m\?rF Ej\2dgF nk\0 ho g\4i h\4{\rs ,\re} \7{m}J\? aAjkl b\7{h}t \7{b}rA lg rhA h\4. m\4{\qva} apn\? pEt aOr uskF yAtnAao\2 s\? Tk \7{c}kF \8{h}\2 .}
    \\(My life has become hell, I feel too bad nowadays. I am tired of my husband and his tortures.)
    \\\\Identification of implicit emotions are sometimes confusing for the annotators due to lack of explicit emotion pointer. For instance, in \textit{Example 2} in which a user is saying that her friend started laughing (in Utterance 3). In the absence of contextual information, this will be perceived as \textit{Joy}. However, by looking at the context of the utterance, this will be annotated with \textit{Surprised,Sad} as emotion labels.
    \\\\\textit{Example 2}: 
    \\Utterance 1: %\begin{sanskrit}मेरे दोस्त को मदद की ज़रूरत है क्योंकि जब उसे पता चला कि उसके पति ने उसे धोखा दिया है तो वह सदमे में है।\end{sanskrit}
    \def\DevnagVersion{2.17}{\dn m\?r\? do-t ko mdd kF j!rt h\4 \3C8wo{\qva}Ek jb us\? ptA clA Ek usk\? pEt n\? us\? DoKA EdyA h\4 to vh sdm\? m\?{\qva} h\4.}
    \\(My friend needs help because she is in trauma after she came to know that her husband cheated on her.)
    \\\\Utterance 2: %\begin{sanskrit}ओह! आपके दोस्त के बारे में सुनकर वाकई दुख हुआ। हमें उसकी मदद करने में हमेशा खुशी होगी। क्या मैं जान सकता हूँ कि वह अब कैसी है?\end{sanskrit}
    \def\DevnagVersion{2.17}{\dn aoh{\rs !\re} aApk\? do-t k\? bAr\? m\?{\qva} \7{s}nkr vAk\4i \7{d}K \7{h}aA. hm\?{\qva} uskF mdd krn\? m\?{\qva} hm\?fA \7{K}fF hogF. \3C8wA m\4{\qva} jAn sktA \8{h}\1 Ek vh ab k\4sF h\4{\rs ?\re}}
    \\(Oh! That's really sad to hear about your friend. We would always be happy to help her. May I know how is she doing now?)
    \\Utterance 3: %\begin{sanskrit}यह सुनते ही वह हंसने लगी।\end{sanskrit}
    \def\DevnagVersion{2.17}{\dn yh \7{s}nt\? hF vh h\2sn\? lgF.}
    \\(She started laughing when she heard this.)
    \item \textbf{Identification of emotions for sarcastic utterances:} Annotating the sarcastic utterances is one of the commonly faced challenges while annotating the utterances of our dataset. Sarcasm is prevalent in most of the previous works in sentiment and emotion analysis. Sarcasm is a sort of verbal irony; simply put, it is something uttered that should be perceived as having the opposite meaning as its literal meaning  \cite{gibbs2007irony}. For instance, in \textit{Example 3}, the emotion closest to the speaker's mood is that of \textit{Anger}, which might easily be misinterpreted as \textit{Joy}. Hence, while annotating the sarcastic utterances, the annotators were instructed to keep in mind the contextual knowledge given by the previous utterances of the dialogue.
    \\\\\textit{Example 3}: %\begin{sanskrit}हा हा हा! अब तुम बताओ मैं क्या करूँ?\end{sanskrit}
    \def\DevnagVersion{2.17}{\dn hA hA hA{\rs !\re} ab \7{t}m btAao m\4{\qva} \3C8wA k!\1{\rs ?\re}}
    \\(Ha ha ha! Now you would tell what should I do?)

\end{itemize}

\begin{figure*}[!ht]
    \centering
    \begin{adjustbox}{max width=0.7\linewidth}
  \includegraphics{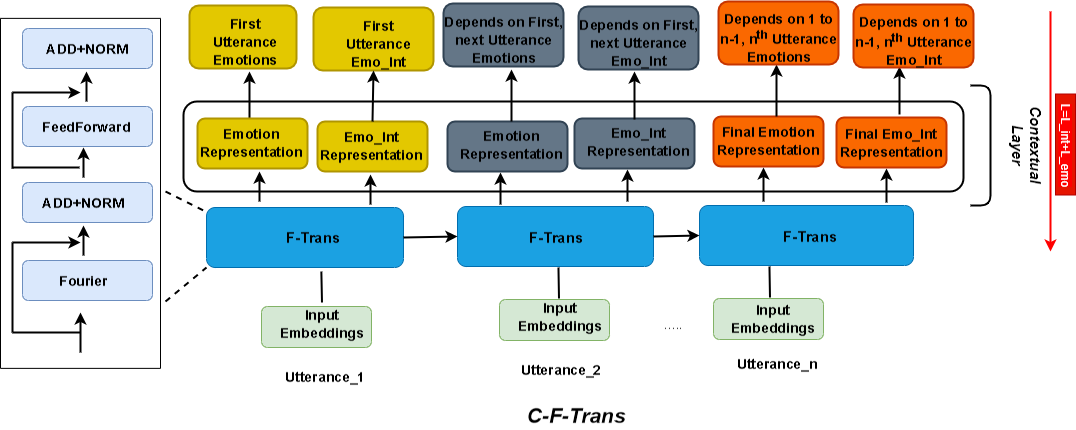}
    \end{adjustbox}
    \caption{Architectural diagram of the C-F-Trans framework}  
    \label{proposed}
\end{figure*}

\subsection{Dataset Statistics}
In Table \ref{datastat}, we provide the important statistics of the dataset followed by the overall emotion distribution of our dataset in Table \ref{emotionstat}.

\subsection{Comparison with the related datasets}
The available datasets for emotion detection are mostly in English. Towards the task of emotion detection from Hindi text, previous attempts have been made in creating corpus containing 2,866 sentences for predicting emotions from Hindi-English code switched language used in social media \cite{vijay2018corpus}. A Hindi dataset, \textbf{IITKGP-SEHSC} consisting of 12,000 sentences collected from auditory speech signals was proposed by \cite{koolagudi2011iitkgp}. Another dataset with 780 Hindi sentences collected from children stories belonging to
three genres, namely fable, folk-tale and legend and annotated with five different emotion categories: happy, sad, anger, fear and neutral was introduced in \cite{harikrishna2016emotion}. Lately, the authors in \cite{kumar2019bhaav} introduced the first largest annotated Hindi corpus named \textbf{BHAAV} for emotion detection consisting of 20,304 sentences from 230 popular Hindi short stories spanning across frequently used 18 genres, \textit{viz.} historical, mystery, patriotic to name a few. The authors in \cite{ahmad2020borrow} proposed Hindi corpus, \textbf{Emo-Dis-HI} consisting of \textit{2,668} sentences from news documents of disaster domain, where each sentence is labeled with one of the emotion categories \textit{viz.}, sadness, sympathy/pensiveness, optimism, fear/anxiety, joy, disgust, anger, surprise and no-emotion. When it comes to the task of analyzing emotions from Hindi text in conversational setting, there are no conversational dataset available in Hindi. Our proposed dataset is different from the existing datasets for emotion detection from Hindi. The dataset that we present here is the first large-scale goal-oriented conversational dataset comprising of 1814 dialogues with each utterance in dialogues annotated for multi-label emotion and corresponding intensity value. Comparisons between the existing datasets and our proposed dataset, EmoInHindi are given in Table \ref{compdata}.

\begin{table}[!ht]
\centering
% \arraystretch{1.3}
\resizebox{0.5\textwidth}{!}
{
\begin{tabular}{|l|c|}
\hline
\bf Parameters & \multicolumn{1}{c|}{\bf CMMEESD}\\  \hline \hline

\multirow{1}{*}{Transformer Encoder Layer}  & 2 \\                          
\multirow{1}{*}{Embeddings}  & 300 \\
\multirow{1}{*}{FC Layer}  & Dropout=0.3 \\ 

\multirow{1}{*}{Activations}  & \textit{ReLu} as activation for our model  \\ 
\multirow{1}{*}{Output}       & Softmax(Emotion, Emotion\_Intensity)  \\ 
\multirow{1}{*}{Optimizer}    & Adam (lr=0.003) \\      
\multirow{1}{*}{Model Loss}   & MultiLabelSoftMarginLoss(Emotion) 
\& negative log-likelihood (Emotion Intensity)\\     
\multirow{1}{*}{Batch}        & 32  \\     
\multirow{1}{*}{Epochs}       & 30  \\ 
\hline    

\end{tabular}
}
\caption{Hyper-parameters for our experiments where.
}
\label{tab:hyperParameters}
\end{table}

% ##################################################
\begin{table*}[!ht]
\centering
\begin{adjustbox}{max width=1.0\linewidth}
% \arraystretch{1.5}
\resizebox{.8\textwidth}{!}
{
% \begin{tabular}{|p{11em}|C{3.5em}|C{4.25em}|C{11em}|} \hline
\begin{tabular}{c|ll|l|l|l|l}
\multirow{2}{*}{\textit{\textbf{METHODS}}} & \multicolumn{2}{c|}{\textit{\textbf{TASK-TYPE}}} & \multicolumn{1}{c|}{\multirow{2}{*}{\textit{\textbf{ACC}}}} & \multicolumn{1}{c|}{\multirow{2}{*}{\textit{\textbf{MICRO-F1}}}} & \multicolumn{1}{c|}{\multirow{2}{*}{\textit{\textbf{HL}}}} & \multicolumn{1}{c}{\multirow{2}{*}{\textit{\textbf{JI}}}} \\ \cline{2-3}
 & \multicolumn{1}{c|}{\textit{\textbf{Emotion}}} & \multicolumn{1}{c|}{\textit{\textbf{Intensity}}} & \multicolumn{1}{c|}{} & \multicolumn{1}{c|}{} & \multicolumn{1}{c|}{} & \multicolumn{1}{c}{} \\ \hline
\multirow{2}{*}{\textit{\textbf{bc-LSTM}}} & \multicolumn{1}{l|}{$\surd$ } & - &0.60 & 0.63 & 0.081 & 0.57   \\
 & \multicolumn{1}{l|}{$\surd$ } &$\surd$   &  0.63 & 0.65 & 0.077 & 0.59  \\ \hline
\multirow{2}{*}{\textit{\textbf{bc-LSTM+ATT}}} & \multicolumn{1}{l|}{$\surd$ } & - &  0.61 & 0.63 & 0.079 & 0.57 \\
 & \multicolumn{1}{l|}{$\surd$ } & $\surd$  & 0.64 & 0.67 & 0.075 & 0.60 \\ \hline
\multirow{2}{*}{\textit{\textbf{CMN}}} & \multicolumn{1}{l|}{$\surd$ } & - & 0.63 & 0.66 & 0.076 & 0.59   \\
 & \multicolumn{1}{l|}{$\surd$ } & $\surd$  & 0.64 & 0.68 & 0.073 & 0.61  \\ \hline
\multirow{2}{*}{\textit{\textbf{C-A-Trans}}} & \multicolumn{1}{l|}{$\surd$ } & - &0.67 & 0.71 & 0.066 & 0.64    \\
 & \multicolumn{1}{l|}{$\surd$ } & $\surd$ & 0.69 & 0.73 & 0.059 & 0.66    \\ \hline
\multirow{2}{*}{\textit{\textbf{C-F-Trans}}} & \multicolumn{1}{l|}{$\surd$ } &- &  0.70 & 0.76 & 0.057 & 0.68  \\
 & \multicolumn{1}{l|}{$\surd$ } &$\surd$  &\textbf{0.72 }& \textbf{0.77} & \textbf{0.055} & \textbf{0.69} 
\end{tabular}
}
\end{adjustbox}
\caption{Results for our proposed framework for Multi-label Emotion  Classification}
\label{res_emoji}
\end{table*}
 
\begin{table*}[ht!]
\small
\centering
\begin{adjustbox}{max width=1.1\linewidth}
\resizebox{1.1\textwidth}{!}
{
% \begin{tabular}{|p{20em}|C{3.5em}|C{4.25em}|C{10em}|} \hline
\begin{tabular}{lllll}
\multicolumn{4}{c}{\textit{\textbf{Correctly Predicted}}} \\ \hline
\multicolumn{1}{l|}{\textit{\textbf{Hindi-Utterance}}} & \multicolumn{1}{l|}{\textit{\textbf{English-Utterance}}} & \multicolumn{1}{l|}{\textit{\textbf{Correct-label}}} & \multicolumn{1}{l|}{\textit{\textbf{Predicted-Label}}} & \textit{\textbf{Predicted Intensity}}\\ \hline
\multicolumn{1}{l|}{%\begin{sanskrit}रक्षक मेरा मकान मालिक मुझे परेशान करने की कोशिश कर रहा है। कृप्या मेरी सहायता करे। \end{sanskrit}
\def\DevnagVersion{2.17}{\dn r\322wk m\?rA mkAn mAElk \7{m}J\? pr\?fAn krn\? kF koEff kr rhA h\4. \9{k}=yA m\?rF shAytA kr\?.}} & 
\multicolumn{1}{l|}{Rakshak my landlord is try to harass me. 
Please help me.} & \multicolumn{1}{l|}{Sad, Annoyed} & \multicolumn{1}{l|}{Sad, \textcolor{red}{Anger}}& 3, 1 \\ \hline
\multicolumn{1}{l|}{%\begin{sanskrit}घिनौना! कोई किसी लड़की के साथ ऐसा कैसे कर सकता है?\end{sanskrit}
\def\DevnagVersion{2.17}{\dn EGnOnA{\rs !\re} koI EksF lXkF k\? sAT e\?sA k\4s\? kr sktA h\4{\rs ?\re}}} & \multicolumn{1}{l|}{Disgusting! How could anyone do this to any girl?} & \multicolumn{1}{l|}{Disgust, Anger} & \multicolumn{1}{l|}{Disgust, Anger} & 3, 2\\ \hline
\multicolumn{1}{l|}{%\begin{sanskrit}तुम मेरा समय क्यों बर्बाद कर रहे हो? अगर आप मेरी मदद नहीं कर सकते तो चले जाओ।\end{sanskrit}
\def\DevnagVersion{2.17}{\dn \7{t}m m\?rA smy \3C8wo{\qva} bbA\0d kr rh\? ho{\rs ?\re} agr aAp m\?rF mdd nhF{\qva} kr skt\? to cl\? jAao.}} & \multicolumn{1}{l|}{Why are you wasting my time? 
If you can't help me go away.} & \multicolumn{1}{l|}{Anger, Annoyed} & \multicolumn{1}{l|}{Anger, Annoyed} & 2, 1\\  \hline
\multicolumn{1}{l|}{%\begin{sanskrit}तुम मेरा समय क्यों बर्बाद कर रहे हो? अगर आप मेरी मदद नहीं कर सकते तो चले जाओ।\end{sanskrit}
\def\DevnagVersion{2.17}{\dn \7{t}m m\?rA smy \3C8wo{\qva} bbA\0d kr rh\? ho{\rs ?\re} agr aAp m\?rF mdd nhF{\qva} kr skt\? to cl\? jAao.}} & \multicolumn{1}{l|}{Why are you wasting my time? If you can't help me go away.} & \multicolumn{1}{l|}{Anger, Annoyed} & \multicolumn{1}{l|}{Anger, Annoyed} & 2, 1\\ \hline

% \multicolumn{4}{c}{\textit{\textbf{Incorrectly Predicted}}} \\ \hline
% \multicolumn{1}{l|}{\textit{\textbf{Hindi-Utterance}}} & \multicolumn{1}{l|}{\textit{\textbf{English-Utterance}}} & \multicolumn{1}{l|}{\textit{\textbf{Correct-label}}} & \multicolumn{1}{l|}{\textit{\textbf{Predicted-label}}}&\textit{\textbf{Predicted-Intensity}} \\ \hline
% \multicolumn{1}{l|}{\sanskrit{मैं आवेदन में क्या लिख ​​सकती हूँ? मुझे प्रिय मत कहो। }} & \multicolumn{1}{l|}{What can I write in the application? Don't call me dear} & \multicolumn{1}{l|}{Annoyed, Anticipation} & \multicolumn{1}{l|}\textcolor{red}{Joy,Grateful} & \textcolor{red}{3,2} 
\end{tabular}
}
\end{adjustbox}
% \caption{Some incorrectly predicted samples by the proposed model.}
\caption{Error analysis: Some correct and incorrectly predicted samples}
\label{tab-error_analysis}
\end{table*}

\section{Baselines}
We use the %give strong benchmarks using the \textit{EmoInHindi} dataset in this section.
following baseline models: %For emotion identification, we use four powerful baseline frameworks. 

\textbf{Baseline \#1: bcLSTM:}
The bidirectional contextual LSTM bcLstm\cite{poria2017context} is a bidirectional contextual LSTM. Two uni-directional LSTMs with opposite directions are stacked to create bi-directional LSTMs. As a result, an utterance may learn from utterances in the video that occur before and after it, which is, of course, context.

\textbf{Baseline \#2: bcLSTM+Attention:}

At each timestamp, an attention module is added to the output of c-LSTM in this bcLSTM with attention\cite{poria2017context} model.

\textbf{Baseline \#3: Conversational Memory Network (CMN):}
Using two different GRUs for two speakers, CMN \cite{hazarika2018conversational} models utterance context from dialogue history. Finally, utterance representation is obtained by querying two separate memory networks for both speakers with the current utterance. However, this model can only model two-person conversations.

% \subsection{Strong Conversation Classification Baseline \#4: DialogueRNN}
% DialogueRNN \cite{majumder2019dialoguernn} is a multi-party framework for moderating emotional and sentimental content in discussions. They described DialogueRNN, a unique recurrent neural network-based system that observes each party's status during a conversation and utilises that information to identify emotions.
% It uses three levels of gated recurrent units (GRU) to capture the conversational environment, allowing it to accurately distinguish emotions, intensity, and attitudes in a debate.
% They demonstrated SOTA's performance on two datasets, IEMOCAP \cite{busso2008iemocap} and AVEC \cite{schuller2012avec}. 

\subsection{ Proposed: C-Attention-Trans and C-Fourier-Trans}
We use the transformer encoder suggested by \cite{vaswani2017attention} for \textit{Context-Attention-Transformer}. A \textit{C-A-Transformer} is proposed to capture the deep contextual relationship with input utterance. %As we all know, contextual utterances can be quite useful in recognising an input utterance. This necessitates a model that considers these inter-dependencies, as well as their potential impact on the target utterance. 
We use a  Transformer-based method to capture the flow of informative triggers across utterances.
The Context-Attention-Transformer receives the input embedding and captures the input utterance's deep contextual relationship.

We employ the Fourier transform instead of self attention, as suggested by F-Net \cite{lee2021fnet} for \textit{Context-Fourier-Transformer}. 1D Fourier Transforms are used to transform both the sequences and hidden dimensions. Instead \textit{self-attention}, this trick proved to be effective. % instead of %Using \textit{Fourier} gives better results instead of using 
%\textit{self-attention}.
\\\\
\textbf{Working method of the model}

Suppose text features have dimension $d$, then each utterance is represented by $u_{i,x}\in \mathcal{R}^d$ where $x$ represents $x^{th}$ utterance of the conversation $i$. %In each conversation, we have utterances. 
To get $U_i$, we collect a number of utterance in a conversation $U_i = [x_{i,1}, x_{i,2}, ..., x_{i,c_i} ] \in \mathcal{R}^{c_i,d}$, where $c_i$ represents the number of utterances we consider as a context in a conversation. This $U_i$ is given to both C-A-Trans and C-F-Trans for the output. We show our model in Fig \ref{proposed}.

The output of the C-A-Transformer and C-F-Transformer is fed to the FC Layer, which then passes it on to the Softmax Layer for emotion and intensity prediction. 

\textbf{Loss function:}
The emotion intensity classifier is trained by minimizing the negative log-likelihood
\begin{equation}
    \mathcal{L}_{Emo} = -\sum_{em=1}^{N} y_{em} \log \tilde{y_{em}}
\end{equation}
 
For multilabel emotion, we use MultiLabelSoftMarginLoss, where $y_{em}$ is the true emotion labels and $\tilde{y_{em}}$ is the predicted emotion label.
For Emotion Intensity, we use MSE (Mean Squared Error) as the loss function.

Our loss function's primary goal is to instruct the model on how to weigh the task-specific losses.
For this, we use a principled approach to multi-task deep learning that takes into account the homoscedastic uncertainty (task dependent or homoscedastic uncertainty is aleatoric uncertainty that is not depending on the input data). Homoscedastic is a number that remains constant throughout all input data and changes between jobs. Task-dependent uncertainty is the effect of this, while weighing multiple loss functions \cite{kendall2018multi} of each task. 
% \begin{equation}
%     \mathcal{L}_{JT} = \mathcal L_{Emo} + \mathcal L_{Gen}
% \end{equation}

\begin{equation}\label{eq2}
   \mathcal L=\sum_{i} \mathcal W_{i}\mathcal L_{i}
\end{equation}

Where $i$ defines different tasks (emotion classification and intensity).

\section{Results and Analysis}
%\textbf{Feature Extraction and Data Distribution for  experiment:}
\subsection{Feature Extraction and Data Distribution for  Experiment}

For textual features, we take the pre-trained 300-dimensional Hindi \textit{fastText} embedding \cite{joulin2016fasttext}. %For which we follow 80:20 ratio for train, test set, and again we divide the train set in to 80:20 ratio in training and validation sets. 
We obtain the training and testing set using 80:20 split of the dataset.
Further, the 20\% of the training set is used as validation set during training to keep
track of model training progress. Empirically, we take five\footnote{Baseline models give the best result at five.} utterances as context for a particular utterance.
 
\subsection{Experimental Setup}
%On the \textit{EmoInHindi} dataset, we assess our suggested model. We use five \footnote{Baseline models produce the greatest results at five.} utterances as the background for a particular utterance. %speech. 

% We use the Python-based PyTorch deep learning library \footnote{urlhttps://pytorch.org/} to develop our suggested model. For Emotion recognition, we use precision (P), recall (R), and F1-score (F1) as assessment metrics. As an optimizer, we employ \textit{Adam}, a classifier for emotion detection called \textit{Softmax}, and a loss function called \textit{categorical cross-entropy}.

%\textbf{Hyperparameter:}
We implement our proposed model in PyTorch, a Python-based deep learning library. We perform \textit{grid search} to find the optimal hyper-parameters in Table \ref{tab:hyperParameters}. We use \textit{Adam} as an optimizer. We use \textit{Softmax} as a classifier for emotion. We use \textit{Transformer Encoder} with two layers. The embedding size is set to 300, and the learning rate is set to %We use learning rate of 
0.003. We use negative log-likelihood loss for emotion prediction. Our model converges with 30 epochs and we use 32 batch size.

\begin{table}[!ht]
\centering
\resizebox{0.5\textwidth}{!}
{    
\begin{tabular}{l|l|l|l|l}
\textbf{Methods} & \textbf{Accuracy} & \textbf{Precision} & \textbf{Recall} & \textbf{F1} \\ \hline
\textit{\textbf{bc-LSTM }}\cite{poria2017context} & 60.95 & 59.13 & 59.72 & 59.42 \\ \hline
\textit{\textbf{bc-LSTM+Attn}}\cite{poria2017context} & 61.91 & 60.54 & 61.82 & 61.18 \\ \hline
\textit{\textbf{CMN}} \cite{hazarika2018conversational}& 63.51 & 63.13 & 63.68 & 63.37 \\ \hline
\textit{\textbf{C-Attention-Trans}} (\textbf{Ours}) & 65.62 & 63.76 & 64.82 & 64.26 \\ \hline
\textit{\textbf{C-Fourier-TRans}} (\textbf{Ours}) & \textbf{67.14} & \textbf{65.98} & \textbf{66.52} & \textbf{66.24}
\end{tabular}
}
\caption{Result Analysis For Single label Emotion with Intensity }
\label{tab:ablation}
\end{table}

% ##################################################

% \input{tex/hyperParameters}

% \footnote{\url{https://pytorch.org/}}

\subsection{Results}

% \subsubsection{with context without speaker: humor classification (MISA+DialogueRNN)}

C-F-Trans, achieves the best precision of 65.98\% ($2.22$ points $\uparrow$ in comparison of C-A-Trans and $2.85$ points $\uparrow$) in comparison of CMN, $5.44$ points $\uparrow$ in comparison of bc-lstm+att, $6.26$ points $\uparrow$ in comparison of bc-lstm, recall of 66.52\% ($1.70$ points $\uparrow$ in comparison of C-A-Trans and $2.84$ points $\uparrow$) in comparison of CMN, $4.7$ points $\uparrow$ in comparison of bc-lstm+att, $6.8$ points $\uparrow$ in comparison of bc-lstm,) and F1-score of 66.24\% ($1.98$ points $\uparrow$ in comparison of C-A-Trans and $2.87$ points $\uparrow$) in comparison of CMN, $5.06$ points $\uparrow$ in comparison of bc-lstm+att, $6.82$ points $\uparrow$ in comparison of bc-lstm,), Accuracy of 67.14\% ($1.52$ points $\uparrow$ in comparison of C-A-Trans and $3.63$ points $\uparrow$) in comparison of CMN, $5.23$ points $\uparrow$ in comparison of bc-lstm+att, $6.19$ points $\uparrow$ in comparison of bc-lstm,). We observe that C-F-Trans performs better than the C-A-Trans. We show the results in Table \ref{tab:ablation} for single label emotion and intensity. 

For Multi-label emotion C-F-Trans, achieves the best HL of 0.055\% ($0.04$ points $\downarrow$ in comparison of C-A-Trans, $0.018$ points $\downarrow$, and $0.020$ points $\downarrow$,in comparison of bc-LSTM+ATT) and C-F-Trans, achieves the best JI of 0.055\% ($0.03$ points $\uparrow$ in comparison of C-A-Trans, $0.08$ points $\uparrow$, and $0.09$ points $\uparrow$ ,in comparison of bc-LSTM+ATT ). We show the results in Table \ref{res_emoji}

\subsection{Error Analysis}
We show a few samples\footnote{For the global audience, we also translate these Hindi utterances into English.} (c.f. Table \ref{tab-error_analysis}) which are correctly predicted by our proposed model (C-F-Trans). For example, as shown in Table \ref{tab-error_analysis}, %\begin{sanskrit} रक्षक मेरा मकान मालिक मुझे परेशान करने की कोशिश कर रहा है। कृप्या मेरी सहायता करे। \end{sanskrit} 
\def\DevnagVersion{2.17}{\dn r\322wk m\?rA mkAn mAElk \7{m}J\? pr\?fAn krn\? kF koEff kr rhA h\4. \9{k}=yA m\?rF shAytA kr\?.} (Rakshak my landlord is try to harass me. Please help me.) have label \textit{Sad,Annoyed} and intensity \textit{3,1}
correctly predicted by our model (C-F-Trans). But our model also confused in some situations as for example, %\begin{sanskrit} मैं आवेदन में क्या लिख सकती हूँ? मुझे प्रिय मत कहो। \end{sanskrit}
\def\DevnagVersion{2.17}{\dn m\4{\qva} aAv\?dn m\?{\qva} \3C8wA ElK sktF \8{h}\1{\rs ?\re} \7{m}J\? E\3FEwy mt kho.} (What can I write in the application? Don't call me dear) have label \textit{Anticipation, Annoyed} but our model predicted \textit{Joy} due to word %\begin{sanskrit} प्रिय\end{sanskrit}
\def\DevnagVersion{2.17}{\dn E\3FEwy} (dear) which is maximum used with emotion \textit{Joy}. So our model predicted \textit{Joy} emotion and \textit{Grateful} as it is come with \textit{Joy} most of the time.%, which is wrong, as the correct emotion is \textit{Anticipation} and \textit{Annoyed}.

\section{Conclusion and Future Direction}
In this paper, we have introduced a large-scale Hindi conversational dataset, \textit{EmoInHindi} prepared in Wizard-of-Oz fashion for multi-label emotion classification and intensity prediction in dialogues. We have evaluated our proposed \textit{EmoInHindi} dataset and reported the results using strong baselines for both tasks of emotion recognition and intensity prediction. We believe that this dataset can be employed in the future for for making emotion-aware conversational agents capable of conversing with the users in Hindi. Furthermore, we would like to extend this work for more low-resource languages like Bengali, Marathi etc. so that it can be used to create emotionally-aware conversational systems that can interact with the users in their regional language thereby creating a more user-friendly environment for them.

%%%%%%%%%%%%%%%%%%%%%%%%%%%%%%%%%%%%%%%%%%%%%%%%%%
\section{Acknowledgement}
Priyanshu Priya acknowledges the Innovation in Science Pursuit for Inspired Research (INSPIRE) Fellowship implemented by the Department of Science and Technology, Ministry of Science and Technology, Government of India for financial support. Asif Ekbal acknowledges the Young Faculty Research Fellowship (YFRF), supported by Visvesvaraya PhD scheme for Electronics and IT, Ministry of Electronics and Information Technology (MeitY), Government of India, being implemented by Digital India Corporation (formerly Media Lab Asia). %All the authors gratefully also acknowledge the partial support from \enquote{IITP Centre of Excellence in Cyber Crime Prevention against Women and Children- AI based Tools for Women and Children Safety}, sponsored by Ministry of Home Affairs, Govt. of India.
\section{Bibliographical References}\label{reference}
%\label{main:ref}

\bibliographystyle{lrec2022-bib}
\bibliography{lrec2022}

% \section{Language Resource References}
% \label{lr:ref}
\bibliographystylelanguageresource{lrec2022-bib}
%\bibliographylanguageresource{languageresource}

\end{document}